\documentclass{article}

\usepackage{spconf,amsmath,graphicx}
\usepackage{multirow}
\usepackage{subcaption}
\usepackage{url}
\usepackage{graphicx}
\usepackage{subcaption}
\usepackage{epstopdf}
\usepackage{xcolor}
\usepackage[bottom]{footmisc}

\title{ Accelerating Proposal Generation  Network for \\Fast Face Detection on Mobile Devices}
\name{Heming Zhang$^1$, Xiaolong Wang$^{2*}$\thanks{* indicates corresponding author}, Jingwen Zhu$^2$, C.-C. Jay Kuo$^1$}
\address{
\begin{tabular}{c c}
$^1$ University of Southern California & $^2$ Samsung Research America\\
Los Angeles, CA, USA & Mountain View, CA, USA\\
\end{tabular}
}

\begin{document}
%
\maketitle
\begin{abstract}
\ninept{Face detection is a widely studied problem over the past few decades. Recently, significant improvements have been achieved via the deep neural network, however, it is still challenging to directly apply these techniques to mobile devices for its limited computational power and memory. In this work, we present a proposal generation acceleration framework for real-time face detection. More specifically, we adopt a popular cascaded convolutional neural network (CNN) as the basis, then apply our acceleration approach on the basic framework to speed up the model inference time. We are motivated by the observation that the computation bottleneck of this framework arises from the proposal generation stage, where each level of the dense image pyramid has to go through the network. In this work, we reduce the number of image pyramid levels by utilizing both global and local facial characteristics (i.e., global face and facial parts). Experimental results on public benchmarks WIDER-face and FDDB demonstrate the satisfactory performance and faster speed compared to the state-of-the-arts. 
}
\end{abstract}
\begin{keywords}
Face detection, mobile devices
\end{keywords}

\section{Introduction}
\ninept{

Face detection has been studied for a long time for its important prerequisite of these face related applications, e.g., face recognition~\cite{schroff2015facenet}, face alignment~\cite{joint_detection_alignment}, face editing~\cite{smith2013exemplar}, face manipulation~\cite{zhang2017age} and tracking~\cite{kim2008face}.  
Early works on face detection mainly rely on hand-crafted features with classifiers. Viola-Jones detector \cite{VJ} is one typical approach which combines the Haar features with AdaBoost classifier. It is still a popular method nowadays due to its small model size and fast speed. Then,  
deformable part models (DPM) based techniques \cite{DPM, DPM2} are becoming popular where latent support vector machine (SVM) is applied to find the parts and their geometric relationship.
Although DPM-based methods have achieved remarkable performance, they are computationally expensive and sensitive to hand-craft features. Recently, a boosted-decision-tree-based face detector \cite{lcdf} outperforms all other non-CNN techniques while operating at the fast speed. However, as discussed in \cite{lcdf}, the detection performance of these boosted trees model is still limited. One major drawback for these approaches is the feature is not learned from the data. This limits the improvement space with additional data and modeling capacity.

With the powerful discriminative capability of the deep neural network, many CNN-based methods have been proposed to solve the face detection problem. One earlier work is proposed by Farfade et al. \cite{yahoo} where a pre-trained AlexNet \cite{alexnet} is used as the basic network structure and converted to a fully-convolutional structure to fit different input face sizes. The feature map is directly used as the heatmap to localize faces. 

Recently, many CNN-based works aim to improve the detection accuracy while compromising model size and speed which can facilitate the mobile device applications. 
Bai et al. \cite{BMVC_multi} add multi-scale branches to the end of the network and reduces the image pyramid to octave-space scales. Yang et al. \cite{wider_face} utilize multiple proposal networks to avoid image pyramid. Recent work \cite{tiny_face} has achieved significant improvements in tiny face detection by training separate detectors and defining multiple templates for different scales.
In \cite{facial_parts}, facial parts heatmaps are obtained from five different networks and combined into a single heatmap. The faceness measure of a candidate bounding box is calculated based on the geometry of each part. The face proposals are then refined by fine-tuned AlexNet \cite{alexnet}. Hao et al. \cite{scale_aware} come with a scale-aware framework where possible sizes of faces are estimated by the scale proposal network. However, the model sizes and computation complexities of these works are still not suitable for mobile devices.

Cascaded CNN face detections \cite{joint_detection_alignment, cascade, joint_cascade, nested_cascade} are favored for its small model size and fast speed compared to other frameworks introduced previously. The cascaded CNN~\cite{cascade} is combined with several shallow networks at different resolutions. However, the network grows larger along with the added cascade. Instead of training each network in the cascade separately, a joint training framework is proposed in \cite{joint_cascade}. To further improve the accuracy of cascaded CNN, face detection and alignment are jointly learned in \cite{joint_detection_alignment}. 
Although these approaches discussed above have relatively small model sizes, the images need to be pre-processed to form the pyramid in order to tolerate various input face sizes. During the inference stage, the input image has to be repeated for each level of the image pyramid, which significantly increases the inference time.

To adapt the CNN-based approaches from high-performance platforms to mobile devices, it is not possible to apply a large and deep structure as discussed above. In this case, we follow the general pipeline as discussed in recent popular cascaded CNN frameworks \cite{cascade}, which consists of several light-weighted networks.
As designed, the first stage of the cascade network serves as a proposal detection stage which quickly scans through the whole image to obtain face candidates. However, the proposal networks can only detect faces within a small range of sizes. For these faces whose sizes exceed the receptive field, it will fail to capture the global facial characteristics. Existing works solve this issue by rescaling the given image to different sizes. The generated images at different scales have to go through the given network which is highly computation inefficient. To speed up the inference procedure, we propose a proposal generation acceleration framework which utilizes both global and local facial cues and enables the multi-scale capability of the proposal stage. For these faces which exceed the processing size, we utilize  local captured facial characteristics as cues to infer the face locations. Consequently, face regions with multiple sizes can be found in a single forward pass. In this way, locating faces of different sizes requires fewer pyramid levels.

The main contributions of this paper can be summarized as:
\begin{itemize}
\item We introduce a new pipeline to accelerate face proposals generation by capturing both global and local facial characteristics. This greatly reduces the number of pyramid levels for the given input image.
\item Our proposed face detector has satisfactory performance and yet meets the crucial memory and speed requirements of mobile devices.
\item Our approach can quickly infer the location of face regions using local facial characteristics instead of relying on the global face.
\end{itemize}}


\section{Proposed Method}
\ninept{\begin{figure}
   \centering  \includegraphics[width=\columnwidth]{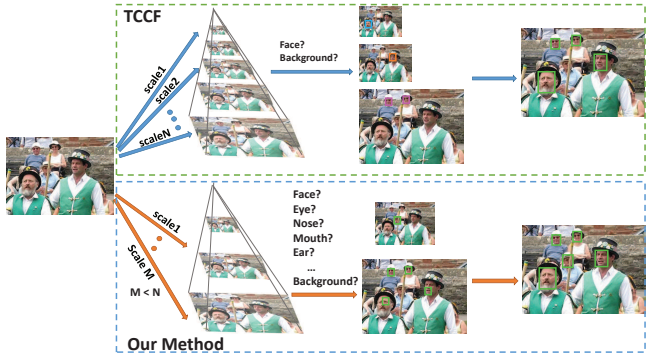}
   \caption{Comparison between our method and typical cascaded CNN frameworks. TCFF indicates the general structure of these typical cascaded CNN frameworks. As we can see, since we also capture local characteristics like eyes, nose, mouth, etc., a single level of the pyramid encodes multiple scales of faces, thus we can reduce the number of pyramid levels and speed up the proposal generation process.}
   \label{fig:compare_with_mtcnn}
\end{figure}

\subsection{Observation and Motivation}
Although the cascaded CNN framework is faster compared to other deep learning structures, it is still not feasible for real-time face detection on mobile devices. This is due to the observation that most of the time is spent on the first stage where it serves as a proposal network and takes each level of the image pyramid as input. Motivated by this observation, we focus on designing a new proposal network to reduce the total amount of image pyramid levels.

In previous cascaded CNN frameworks, the image needs to be resampled to the right size to make sure the face region matches to the receptive field of the proposal network (12$\times$12 is used in \cite{joint_detection_alignment, cascade, joint_cascade,  nested_cascade}).  Each pyramid level corresponds to a specific scale of the face. As a result, we need to form a dense image pyramid in order to achieve high detection accuracy. One way to reduce the computation time is to directly reduce the number of pyramid levels. However, the accuracy will drop rapidly. If we can encode more scales per pyramid level, less sparse pyramid levels will be needed, and then the proposal generation process will be accelerated. 

Based on this motivation, we propose a novel proposal module that not only focuses on the global characteristics of the face, but also captures some local cues. If the global characteristic is captured, the input patch will be directly passed to the next stage as a face proposal. On the other hand, when local cues are captured, the location of the face is inferred and the corresponding region is used as the face proposal. A comparison between our method and previous cascaded CNN frameworks is demonstrated in Fig.\ref{fig:compare_with_mtcnn}. Our proposed proposal acceleration pipeline is illustrated in Fig. 
\ref{fig:scheme}.

\begin{figure*}
   \centering
   \includegraphics[width=1.7\columnwidth, trim={0 5cm 0 7.5cm}]{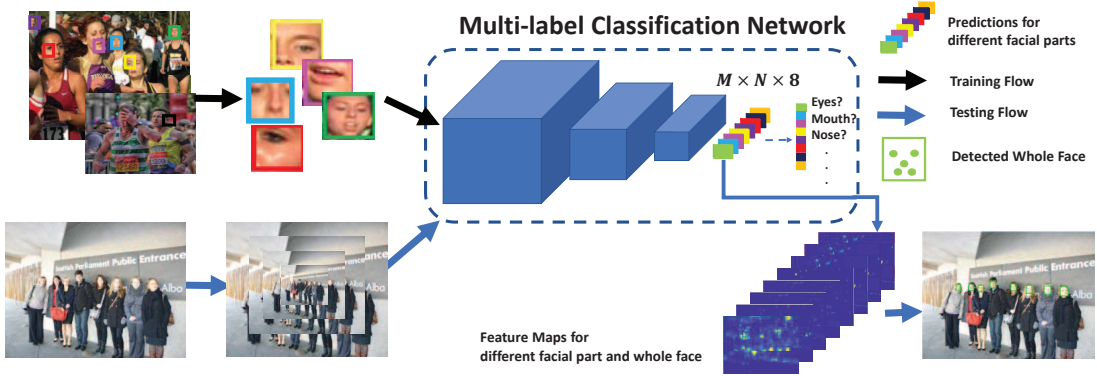}
   \caption{An illustration of the proposal module. In the training stage, face and facial part patches are randomly cropped from training images, and used to train a multi-label classification network. In the testing stage, a test image is resized to form a sparse pyramid, and fed into the multi-label classification network to generate heatmaps of face and facial parts. Based on the heatmaps and bounding box templates of each facial parts, we can generate face proposals. These face proposals will be sent to the next stage of the cascaded framework.}
   \label{fig:scheme}
\end{figure*}

\subsection{Proposal Network Design}
Our proposal network needs to be able to capture both global and local characteristics of faces. Therefore we design the network as a multi-label classifier which can classify an input patch to the background, global face or other facial parts. Since we do not want to confuse the classifier with similar regions such as cheek and forehead, we only choose the most distinctive facial parts such as eye, nose and mouth.

Table. \ref{tb:network} gives the details of our proposal network for training. The input resolution of the proposal network is 12$\times$12, the same as previous work \cite{joint_detection_alignment, cascade, joint_cascade, nested_cascade}. Inspired by \cite{joint_detection_alignment} and \cite{yahoo}, we design the network to be fully-convolutional. Therefore we can directly apply the network to images with arbitrary dimension and avoid cropping out patches using sliding window. The stride of the whole network is set to 2. When it scans through an input image during testing, it is equivalent to using sliding window of stride 2.

\begin{table}
	\centering
	\caption{Proposal network architecture}
		\label{tb:network}
		\begin{tabular}{|c|c|c|}
			\hline
			Layer & Kernel size & Output size\\
			\hline
			Input & & 12$\times$12$\times$3 \\
			\hline
			Conv1 & 3$\times$3 & 12$\times$12$\times$16 \\ 
			\hline
			Pool1 & 3$\times$3 & 6$\times$6$\times$16 \\ 
			\hline
			Conv2 & 3$\times$3 & 4$\times$4$\times$32 \\ 
			\hline
			Conv3 & 3$\times$3 & 2$\times$2$\times$32 \\ 
			\hline
			Conv4 & 2$\times$2 & 1$\times$1$\times$64 \\ 
			\hline
			Conv5 & 1$\times$1 & 1$\times$1$\times$8 \\ 
			\hline
		\end{tabular}
\end{table}

%

\subsection{Proposal Generation}
During the detection, given heatmaps of the global face and facial parts, we need to generate proposals in terms of bounding boxes. The bounding boxes generation from face heatmap and facial part heatmaps are processed separately. For face heatmap, similar to \cite{yahoo}, a threshold $\tau_f$ is applied to the heatmap and local maximums on the heatmap are extracted to generate bounding boxes. 

Fast proposal generation from parts is a non-trivial problem. Previous approaches relying on facial parts \cite{DPM, facial_parts, 2017FG_encoder_decoder} are computational intensive when combining facial part region with the global face via sliding windows. 
Unlike these works, our proposed method aims at utilizing facial parts to reduce the number of image pyramid levels and speed up the detection process. Since the number of faces is much less than the number of sliding windows, it is time-consuming to evaluate each sliding window. Furthermore, we do not want to spend extra time on generating generic object proposals. Therefore instead of using candidate window approach and scoring each window, we propose to directly generate candidate bounding boxes from our facial part heatmaps. These bounding boxes are then combined and evaluated simultaneously. Our pipeline consists of three steps:


\begin{enumerate}
\item \textbf{Finding local maxima} \\
For each facial part heatmap, we first apply threshold $\tau_p$ to find the strong response, where $p$ denotes a certain facial part. Non-maximum suppression (NMS) is then applied to obtain the strongest response points in local regions of heatmap.

\item \textbf{Bounding box generating using templates} \\
We define bounding box template(s) of the face for each facial part. For eyes, we define two templates since we do not identify left eyes and right eyes. Each bounding box is determined by  coordinates of its upper-left vertex $(x_1, y_1)$ and bottom-right vertex $(x_2, y_2)$. We denote a bounding box $i$'s location as $\mathbf{b}_i=(x_{i1},y_{i1},x_{i2}, y_{i2})$. For bounding box $i$, we define its score $p_i$ as the corresponding value on the heatmap. In this way, we can roughly sketch the bounding boxes of the face based on detected local maximums from the previous step.

\item \textbf{Part box combination}\\
For the bounding boxes generated from different facial parts, we employ a similar way as NMS to combine them, which is described as follows.

Given a set of bounding boxes, we start from the bounding box with highest score and find all bounding boxes that have intersection over union (IoU) with it higher than threshold $\tau_{IoU}$. By taking the average of their coordinates, those bounding boxes are merged together:
\begin{equation}
\begin{split}
&\mathbf{b}_{m,i} = \frac{1}{|\mathcal{C}_i|} \sum_{j\in \mathcal{C}_i} \mathbf{b}_j,\\
\text{where }&\mathcal{C}_i = \{\mathbf{b}_i\} \bigcup \{\mathbf{b}_j: IoU(\mathbf{b}_i, \mathbf{b}_j) > \tau_{IoU} \}.
\end{split}
\label{eq:loc}
\end{equation}
The score of the merged bounding box is defined as
\begin{equation}
p_{m,i} = 1-\prod_{j \in \mathcal{C}_i} (1-p_j),
\label{eq:prob}
\end{equation}
which resembles the statistic rule of combining two independent events.
The merged bounding box is assigned to the proposal set and the bounding boxes used for merging are eliminated from the original set. Then we repeat the searching and merging process for the remaining bounding boxes in the original set. This process is repeated until there is no remaining bounding boxes left.

\end{enumerate}}


\section{Experiments}
\ninept{\subsection{Experimental Setup}

As stated before, our accelerating proposal module can be combined with any face classifiers. To train a small model with satisfactory performance, we cascade it with a CNN to construct the whole face detection pipeline. Specifically, after the proposal module, we adopt two successive sub-networks that follow the same structure as the RNet (second stage) and the ONet (last stage) used in the MTCNN
\cite{joint_detection_alignment}. As a result, we form a three-stage cascaded lightweight deep face detector. 

We evaluate the proposed face detector on two popular benchmarks: the WIDER-face \cite{wider_face} and the FDDB \cite{fddb}.  The WIDER-face dataset has 393,703 labeled face bounding boxes from 32,203 images while the FDDB dataset contains 5,171 annotated faces.  To build our training dataset, we use the WIDER-face \cite{wider_face} training set to extract background and face patches. The WIDER-face dataset consists of 32,000 images, where 50\% of them are used for testing, 40\% for training and the remaining
ones are for validation.  Furthermore, the eye, nose and mouth patches for training are extracted from the CelebA \cite{celeba}, which has around 200,000 images and most of the images contain a single face with landmark locations provided. 


\subsection{Evaluation of Model Size}

We compare the size of our model with other works. The results are listed in Table. \ref{tb:size}, where * denotes our calculation based on the information from the literature, and the rest is directly measured. From the comparison, our model is much smaller compared to those complicated CNN frameworks such as DDFD \cite{yahoo}, HR \cite{tiny_face}, CEDN \cite{2017FG_encoder_decoder}. It is even smaller
than LCDF+ \cite{lcdf}, which is the state-of-the-art method that uses manually-crafted feature framework. Our framework also has comparable size to other cascaded CNN-based models (MTCNN \cite{joint_detection_alignment}, nested CNN detector\cite{nested_cascade}). 

\begin{table}
\caption{Comparisons of model size with state-of-the-art networks.}
\label{tb:size}
\begin{center}
\begin{tabular}{|c|c|}
\hline
Work & Model size\\
\hline
CEDN \cite{2017FG_encoder_decoder} & 1.1GB* \\
\hline
DDFD \cite{yahoo} & 233MB* \\
\hline
HR \cite{tiny_face} & 98.9MB\\
\hline
LCDF+ \cite{lcdf} & 2.33MB \\
\hline
MTCNN \cite{joint_detection_alignment} & 1.9MB \\
\hline
Nested \cite{nested_cascade} & 1.6MB* \\
\hline
Ours & 1.96MB\\
\hline
\end{tabular}
\end{center}
\end{table}

\subsection{Evaluation of Face Detection \label{eval_FD}}

\noindent{\bf Multi-scale capability}

We compare the multi-scale capabilities of our face detector and the MTCNN \cite{joint_detection_alignment}, which is the state-of-the-art cascaded CNN face detection engine. The WIDER-face validation set \cite{wider_face} with large face scale
variety is used for evaluation. . It consists of three subset, namely the easy, medium and hard sets. Since this experiment targets at evaluating the performance in multi-scale detectability, we set different levels of scaling factor for the image pyramid. For fair comparison, we use the model provided and follow the same parameter setting in \cite{joint_detection_alignment}. The results are listed in Table \ref{tb:comp}. 

From the results, we can find that both face detectors achieve satisfactory accuracy with the dense image pyramid at the scale factor of 0.79. This scale is also chosen by MTCNN. Our detector outperforms the MTCNN on the Easy set, while the MTCNN performs better on the Hard set. It is worth noting that the MTCNN utilizes joint training for face detection and facial landmark localization. The latter is not used in our detector training. 

As the image pyramid becomes sparse, MTCNN's accuracy drops rapidly. When the scale factor decreases from 0.79 to 0.25, its accuracy degrades by 8.1\%, 6.5\% and 9.3\% on the easy, medium and hard sets, respectively. In contrast, the accuracy of our method without model acceleration drops by 1.8\%, 1.5\%, 8.4\%, respectively.

\begin{table}
\caption{Comparisons of detection performance with MTCNN\cite{joint_detection_alignment} on WIDER-face validation set \cite{wider_face} with different scale factors. }
\label{tb:comp}
\begin{center}
\begin{tabular}{|c|c|c|c|c|}
\hline
\multicolumn{2}{|c|}{Scale factor} & 0.79 & 0.50 & 0.25 \\
\hline
\multirow{2}{*}{Easy} & MTCNN \cite{joint_detection_alignment}& 0.836 & 0.817 & 0.755\\
\cline{2-5}
& Ours & 0.844 & 0.842 & 0.826\\
\hline
\multirow{2}{*}{Medium} & MTCNN \cite{joint_detection_alignment}& 0.809 & 0.798 & 0.744\\
\cline{2-5}
& Ours & 0.809 & 0.805 & 0.794\\
\hline
\multirow{2}{*}{Hard} & MTCNN \cite{joint_detection_alignment}& 0.622 & 0.600 & 0.529\\
\cline{2-5}
& Ours & 0.603 & 0.568 & 0.519\\
\hline
\end{tabular}
\end{center}
\end{table}

\noindent{\bf Accuracy benchmarks} 

We conduct face detection experiments on the FDDB \cite{fddb}. We use 0.25 as the pyramid scaling factor and add an extra layer to the image pyramid with half of the size of the largest scale. As illustrated in Fig. \ref{fig:fddb}, our method outperforms many others such as the CEDN \cite{2017FG_encoder_decoder} and the nested CNN detector \cite{nested_cascade}. In terms of detection accuracy, our model can achieve 94.35\%. As compared to 83.29\% obtained by the nested CNN detector, we have 9.2\% improvement. It also achieves comparable accuracy as compared to the MTCNN \cite{joint_detection_alignment} that uses more pyramid levels.  The HR \cite{tiny_face} outperforms our model by a small margin, yet its model size is too large to be deployed on mobile devices. 

\begin{figure}
\centering
\includegraphics[width=\columnwidth]{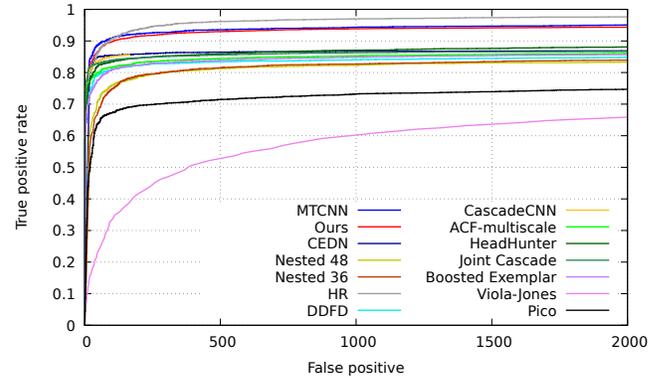}
\caption{Evaluation results on FDDB.}\label{fig:fddb}
\end{figure}

\subsection{Evaluation of Runtime Efficiency}

\noindent{\bf Runtime comparison with MTCNN \cite{joint_detection_alignment}}

We compare the detection speed with the MTCNN method using their provided Matlab codes. For fair comparison, our method is also implemented in Matlab codes. The experiment was conducted on the
WIDER-face validation set \cite{wider_face} using the GeForce GTX TITAN. We use original images without re-sampling to a fixed resolution. The runtime is calculated by averaging the time over the entire validation set. For both detectors, the minimum face size to detect is set to 10 as used by the MTCNN. The scaling factor of the MTCNN is 0.79 as given in its original setting, while scaling factor of 0.25 and an extra pyramid layer is the setting for our detector with comparable accuracy listed in Sec. \ref{eval_FD}. For some images in the WIDER-face validation set, the number of face proposals generated by the MTCNN is more than that our GPU memory (12GiB) can take. Therefore, we take at most 20000 proposals per image for the MTCNN, which in fact reduces the average runtime of the MTCNN.  The average runtime of the MTCNN in this case is 0.595s while that of our detector is 0.499s. We reach more than 16\% acceleration.  Clearly, our detector achieves comparable accuracy with a faster speed. 

\noindent{\bf Runtime comparison with Nested CNN Detector \cite{nested_cascade}}

The running time claimed by the nested CNN detector \cite{nested_cascade} is 40.1ms using the CPU only, where 640$\times$480 VGA image with 80$\times$80 as the minimum size. For comparison, we follow the same setting of the resolution and the minimum face size. The data used for runtime evaluation was not mentioned in \cite{nested_cascade}. Here, we evaluate detection accuracy and running time on FDDB \cite{fddb} with the same dataset in \cite{nested_cascade} for performance benchmarking. With model acceleration, our model can get 39.1ms compared to 40.1ms achieved by the nested CNN detector. It shows
that we can still get a faster s peed with significant accuracy improvement as indicated in Sec. \ref{eval_FD}. 

\noindent{\bf Speed on mobile devices}

We implemented our method on Samsung Galaxy S8 using Caffe. The face detector received images of high resolution (1280$\times$720) from the back camera continuously. By setting the minimum face size to 100 and scaling factor to 0.25, even with the extra pyramid layer mentioned in previous experiments, the detection speed still achieves 8 to 10 FPS in different scenarios on mobile CPU.

}

\section{Conclusion}
\ninept{In this paper, we presented an efficient face detector. Particularly, we proposed a new framework to quickly generate face proposals by capturing both global and local facial cues to reduce image pyramid levels and introduced a method to infer face locations from local facial characteristics.  We validated the proposed methods on two popular benchmarks. The promising performance over  the  state-of-the-art in terms of accuracy, model size and detection speed   demonstrates the potential of our approach towards the real deployment on mobile devices.


}


\bibliographystyle{IEEEbib}
\bibliography{ref}

\end{document}